\documentclass[10pt]{article}
\usepackage{graphicx}
\usepackage{geometry} 
\usepackage{color}
\raggedright
\makeatletter
\renewcommand{\@biblabel}[1]{\quad#1.}
\makeatother

\date{}

\begin{document}
\title{Neutral evolution and turnover over centuries of English word popularity}
\maketitle

\begin{flushleft}
{\Large
\textbf\newline{} 
}
\newline
\\
Damian Ruck\textsuperscript{1,2,3},
R. Alexander Bentley\textsuperscript{2,3,*},
Alberto Acerbi\textsuperscript{4},
Philip Garnett\textsuperscript{5},
Daniel J. Hruschka\textsuperscript{6},
\\
\bigskip
\textbf{1} Bristol Centre for Complexity Sciences, University of Bristol, UK
\\
\textbf{2} School of Social and Community Medicine, University of Bristol, UK
\\
\textbf{3} Hobby School of Public Affairs, University of Houston, USA
\\
\textbf{4} Eindhoven University of Technology, Netherlands
\\
\textbf{5} York Management School, University of York, UK
\\
\textbf{6} School of Human Evolution and Social Change, Arizona State University, USA
\\
\bigskip

* rabentley@uh.edu

\end{flushleft}
{\bf Here we test Neutral models against the evolution of English word frequency and vocabulary at the population scale, as recorded in annual word frequencies from three centuries of English language books.  Against these data, we test both static and dynamic predictions of two neutral models, including the relation between corpus size and vocabulary size, frequency distributions, and turnover within those frequency distributions. Although a commonly used Neutral model fails to replicate all these emergent properties at once, we find that modified two-stage Neutral model does replicate the static and dynamic properties of the corpus data. This two-stage model is meant to represent a relatively small corpus (population) of English books, analogous to a `canon', sampled by an exponentially increasing corpus of books in the wider population of authors. More broadly, this model\textemdash a smaller neutral model within a larger neutral model\textemdash  could represent more broadly those situations where mass attention is focused on a small subset of the cultural variants.}

\section*{Introduction}
English has evolved continually over the centuries, in the branching off from antecedent languages in Indo-European prehistory \cite{Lieberman_etal_2007, Pagel_etal_2007}, in the rates of regularisation of verbs \cite{Lieberman_etal_2007} and in the waxing and waning in the popularity of individual words \cite{Altman_etal_2011, Bentley_2008, Michel_etal_2011}. At a much finer scale of time and population, languages change through modifications and errors in the learning process \cite{Christiansen&Chater_2008, Hruschka_etal_2009}. 

This continual change and diversity contrasts with the simplicity and consistency of Zipf's law, by which the frequency a word, $f$, is inversely proportional to its rank $k$, as $f \sim k^{- \gamma}$ and Heaps law, by which vocabulary size scales sub-linearly with total number of words, across diverse textual and spoken samples \cite{Li_1992, Perc_2012, Sigurd_etal_2004, Zipf_1949, Clauset_etal_2009, Gabaix_2009, Williams_etal_2015, Petersen_etal_2012}.  

The Google Ngram corpus \cite{Michel_etal_2011} provides new support for these statistical regularities in word frequency dynamics at timescales from decades to centuries \cite{Gao_etal_2012, Perc_2012, Petersen_etal_2012, Acerbi_etal_2013, Hughes_etal_2012}.  With annual counts of n-grams \textemdash  an n-gram being $n$ consecutive character strings, separated by spaces \textemdash derived from millions of books over multiple centuries \cite{Lin_etal_2012}, the n-gram data now covers English books from the year 1500 to year 2008. 

In English, the Zipf's law in the n-gram data \cite{Perc_2012} exhibits two regimes: one among words with frequencies above about $0.01\%$ (Zipf's exponent $\gamma \approx 1$) and another ($\gamma \approx 1.4$) among words with frequency below $0.0001\%$ \cite{Petersen_etal_2012}. The latter Zipf's law exponent $\gamma$ of 1.4 is equivalent to a probability distribution function (PDF) exponent, $\alpha$, of about 1.7 ($\alpha = 1+1/\gamma$).    

In addition to the well-known Zipf's law, word frequency data have at least two other statistical properties. One, known as Heaps law, refers to the way that vocabulary size scales sub-linearly with corpus size (raw word count). The n-gram data show Heaps law in that, if $N_t$ is corpus size and $v_t$ is vocabulary size at time $t$, then $v_t \approx N_t^\beta$, with $\beta \approx 0.5$, for all English words in the corpus \cite{Petersen_etal_2012}. If the n-gram corpus is truncated by a minimum word count, then as that minimum is raised the Heaps scaling exponent increases from $\beta < 0.5$, approaching $\beta < 1$ \cite{Petersen_etal_2012}. 
 
The other statistical property is dynamic turnover in the ranked list of most commonly used words.  This can be measured in terms of how many words are replaced through time on ``Top $y$'' ranked lists of different sizes $y$ of most frequently-used words \cite{Bentley_etal_2007, Eriksson_etal_2010, Evans&Giometto_2012, Ghoshal&Barabasi_2011}. We can define this turnover $z_y(t)$ as the number of new words to have entered the top $y$ most common words in year $t$, which is equivalent to the  the top $y$ in that year. The plotting of turnover $z_y$ for different list sizes $y$ can therefore be useful in characterising turnover dynamics \cite{Acerbi_Bentley_2014}.

Many functional or network models readily yield the static Zipf distribution \cite{Gabaix_2009, Clauset_etal_2009} and Heaps law \cite{Lu_etal_2010}, but not the dynamic aspects such as turnover. Here we focus on how Heaps law and Zipf's law can be modeled together with continual turnover of words within the rankings by frequency \cite{Batty_2006, Ghoshal&Barabasi_2011}. We focus on the 1-grams in Google's English 2012 data set, which samples English language books published in any country \cite{Google_2016}. 

\section*{Neutral models of vocabulary change}
One promising, parsimonious approach incorporates the class of neutral evolutionary models \cite{Bentley_etal_2004, Bentley_etal_2007, Bentley_etal_2014b, Gleeson_etal_2014, Neiman_1995} that are now proving insightful for language transmission \cite{Bentley_2008, Bentley_etal_2011b, Reali&Griffiths_2009}. The null hypothesis of a Neutral model is that copying is undirected, without biases or different `fitnesses' of the words being replicated \cite{Acerbi_Bentley_2014, Kandler&Shennan_2013}. 

\begin{figure}[!h]
  \begin{center}
\includegraphics[width=2.3in]{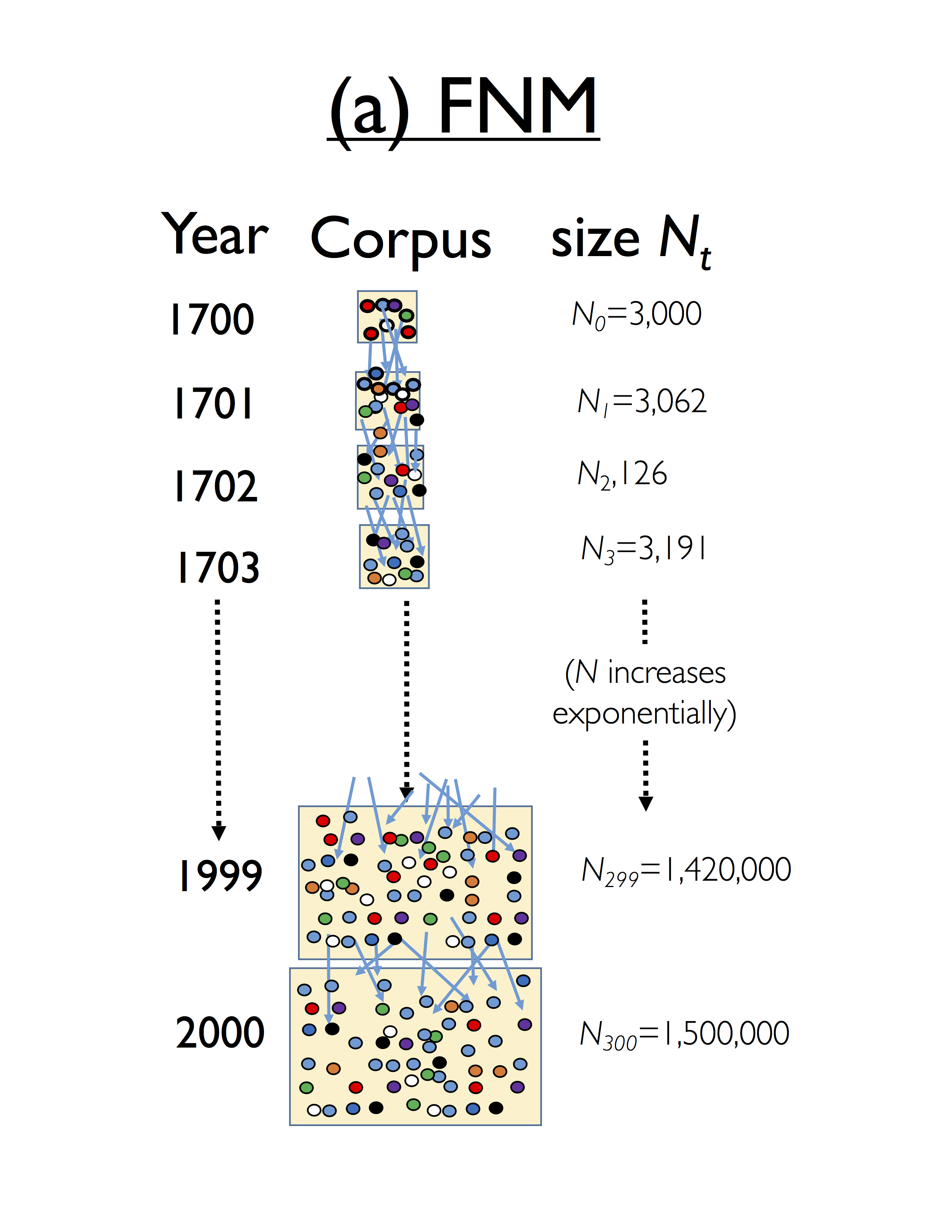}
 \includegraphics[width=2.3in]{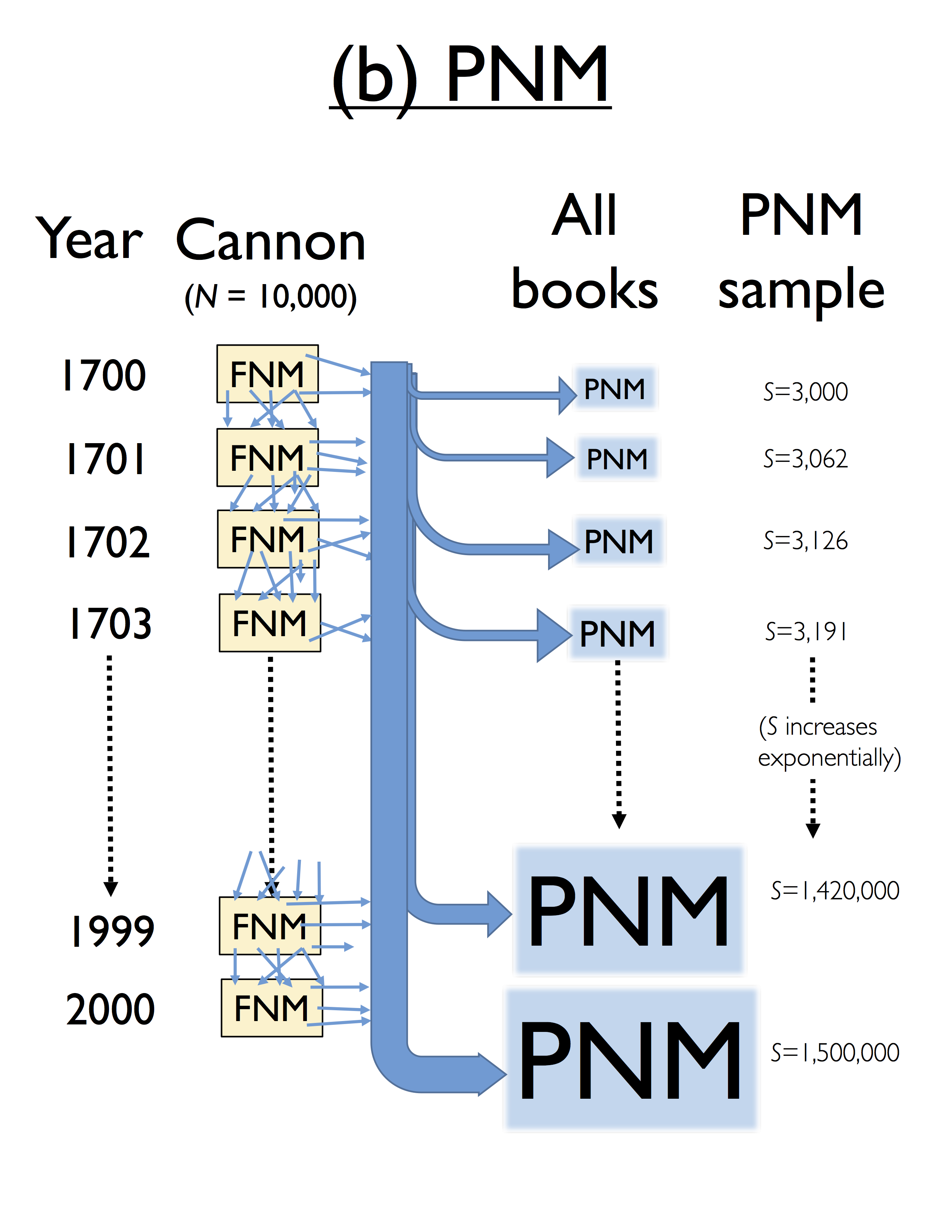}
  \end{center}
  \caption{{\bf Schematic representation of the Full-sampling Neutral model (FNM) and Partial sampling Neutral model (PNM).  (a)} In the FNM, each of the $N_t$ words in year $t$, represented by different colored circles in each box, is copied (arrows) from from the previous year $t-1$ with probability $1-\mu$, or newly-invented with probability $\mu$. The FNM shown in (a) has a corpus size $N_t$ that grows through time. In {\bf (b)} the PNM samples from all previous results of the FNM since the initial time step representing year 1700. The PNM population grows exponentially ($N_0e^{0.021t}$) through time, from 3000 to 1.5 million. As the PNM samples from all previous years of FNM population, the PNM samples from a corpus that increases linearly (by 10,000 words per year) from 10,000 words in year 1700 to 3 million words by year 2000. For the  PNM, the big blue arrows represent how each generation can sample any year of the canon randomly, all the way back to 1700, the smaller arrows representing individual sampling events.}
 \label{Models} 
 \end{figure}

A basic neutral model, which we will call the {\it full-sampling Neutral model} (FNM), would assume simply that authors choose to write words by copying those published in the past and occasionally inventing or introducing new words. As shown in Fig~\ref{Models}a, the FNM represents each word choice by an author as selecting at random among the $N_t$ words that were published in the previous year \cite{Reali&Griffiths_2009, Bentley_etal_2011b}. This copying occurs with probability $1-\mu$, where $\mu \ll 1$ is the fixed, dimensionless probability that an author invents a new word (even the word had originated somewhere `outside' books, e.g. in spoken slang). Each newly-invented word enters with frequency one, regardless of $N_t$. In terms of the modeled corpus, a total of about $\mu N_t$ unique new words are invented per time step. Note that $N_t$ represents the total number of written words, or corpus size, for year $t$, which contrasts with the smaller ``vocabulary'' size, $v_t$, defined as the number of {\it different} words in each year $t$ regardless of their frequency of usage. 

As has been well demonstrated, the FNM readily yields Zipf's law \cite{Bentley_etal_2004, Bentley_etal_2011a, Strimling_etal_2009}, which can also be shown analytically (see Appendix 1). Also, simulations of the FNM show that the resulting Zipf distribution undergoes dynamic turnover \cite{Bentley_etal_2007}.  Extensive simulations \cite{Evans&Giometto_2012} show that when list size $y$ is small compared to the corpus ($0.15y<N_t\mu$), this neutral turnover $z_y$ per time step is more precisely approximated by:

\begin{equation}
	z_y=1.4\cdot \mu^{0.55}\cdot y^{0.86}\cdot n^{0.13},
\label{EG1}
\end{equation}

\noindent where $n$ is the number of words per time interval. 

This prediction can be visualized by plotting the measured turnover $z_y$ for different list sizes $y$. The FNM predicts the results to follow $z_y \propto y^{0.86}$, such that departures from this expected curve can be identified to indicate biases such as conformity or anti-conformity\cite{Acerbi_Bentley_2014}.  It would appear from eq. \ref{EG1} that turnover should increase with corpus size. This is the nominal equilibrium for FNM with constant $N_t$. If corpus size $N_t$ in the FNM is growing exponentially with time, however, then there may be no such nominal equilibrium. In this case we predict that the turnover $z_y$ can actually decrease with time as $N_t$ increases. This is because newly invented words start with frequency one, and under the neutral model they must essentially make a stochastic walk into the top 100, say. As $N_t$ grows, so does the minimum frequency needed to break into the top 100. As the ``bar'' is raised, words are more likely to `die' before they ever reach the bar by stochastic walk \cite{Petersen_etal_2012b}. As a result, turnover in the Top $y$ can slow down over time and growth of $N_t$. 

The FNM does not, however, readily yield Heaps law ($v_t=N_t^\beta$, where $\beta < 1$), for which $\beta \approx 0.5$ among the 1-gram data for English \cite{Petersen_etal_2012}. In the FNM, the expected exponent $\beta$ is 1.0, as the number of different variants (vocabulary) normally scales linearly with $\mu N_t$  \cite{Bentley_etal_2004}. 

While the FNM has been a powerful null model, in the case of books, we can make a notable improvement to account for the fact that most published material goes unnoticed while a relatively small portion of the corpus is highly visible. To name a few examples across the centuries, literally billions of copies of the Bible and the works of Shakespeare have been read since the seventeenth century, as well as tens or hundreds of millions of copies of works by Voltaire, Swift, Austen, Dickens, Tolkien, Fleming, Rawling and so on. While these and hundreds more books become considered part of the ``Western Canon,'' that canon is constantly evolving \cite{Hughes_etal_2012} and many books that were enormously popular in their time \textemdash {\it e.g., Arabian Nights} or the works of Fanny Burney\textemdash fall out of favour. As the published corpus has grown exponentially over the centuries, early authors were more able to sample the full range of historically published works, whereas contemporary authors sample from an increasingly small and more recent fraction of the corpus, simply due to its exponential expansion \cite{Hughes_etal_2012, Pan_etal_2016}.

As a simple way of capturing this, we propose a modified neutral model, called the {\it partial-sampling Neutral model} (PNM), of an evolving ``canon'' that is sampled by an exponentially-growing corpus of books. As shown in Fig~\ref{Models}b, the PNM represents an exponentially growing number of books that sample words from a fixed size canon over all previous years since 1700. Our PNM represents a world where there exists an evolving canonical literature as a relatively small subset of the world's books on which all writers are educated. As new contributions to the canon are contributed, authors sample from the recent generation of writers with occasional innovation. Because the canon is a high-visibility subset of all books, only a fixed, constant number words of text per year are allowed into a year's canon. The rest of the population learns from the cumulative canon since our chosen reference year of 1700.

 \section*{Results}
The average result from 100 runs in each of the FNM and PNM  were used to match summary statistics with the 1-gram data.  Several key statistical results emerge from analysis of the 1-gram data which we compare the FNM to the PNM in terms of these results: (1) Heaps law, which is the sublinear scaling of vocabulary size with corpus size, (2) a Zipf's law frequency distribution for unique words, (3) a rate of turnover that decreases exponentially with time and a turnover vs popular list size that is approximately linear. Here we describe our results in terms of rank-frequency distributions, turnover and corpus and vocabulary size.  We compare the partial-sample Neutral model (PNM) to the full 1-gram data for English. 

\begin{figure}[!h]
  \begin{center}
       \includegraphics[width=2.3in]{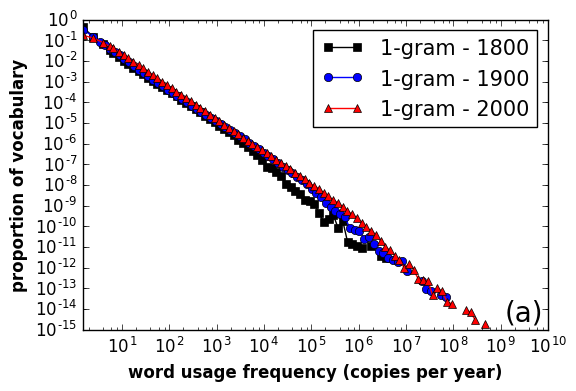}
      \includegraphics[width=2.3in]{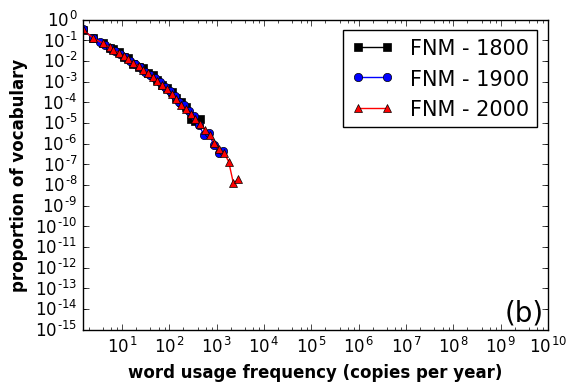}
   \includegraphics[width=2.3in]{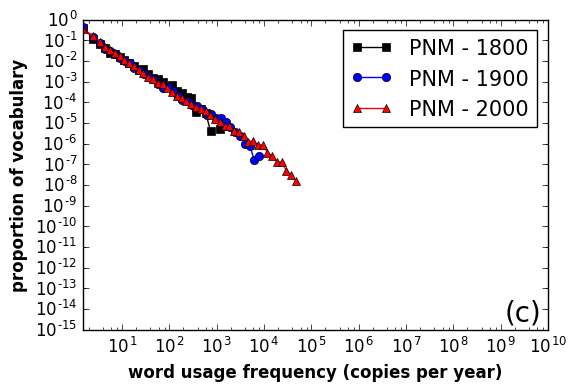}
    \includegraphics[width=2.3in]{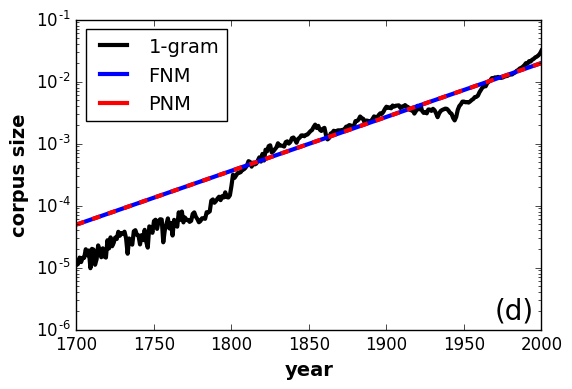}
  \end{center}
  \caption{{\bf Rank-frequency distributions among English words, (a)} In the 1-gram corpus. Black symbols show the distribution for the year 1800, blue shows year 1900 and red shows year 2000. The simulated results are shown for the FNM in {\bf (b)} and the PNM in {\bf (c)}. Panel {\bf (d)} shows the actual number of English words, $N_t$ in the 1-gram corpus versus the modeled corpus size $N_0e^{0.021t}$, where $t$ is number of years since1700.}
 \label{Zipf} 
 \end{figure}

First, we check that the model replicates the Zipf's law that characterizes the 1-gram frequencies in multiple languages \cite{Perc_2012}. Our own maximum likelihood determinations, applying available code \cite{Clauset_etal_2009} to the Google 1-gram data, confirm that the mean $\alpha = 1.75 \pm 0.12$ for the Zipf's law over all English words in the hundred years from 1700 to 1800 (beyond 1800, the corpus size becomes too large for our computation). Normalising by the word count \cite{Gabaix_2009}, the form of the Zipf distribution is virtually identical for each year of the dataset, reaching eight orders of magnitude by the year 2000 (Fig~\ref{Zipf}a). The FNM replicates the Zipf (Fig~\ref{Zipf}b) but the PNM replicates it better and over more orders of magnitude (Fig~\ref{Zipf}c). It was not computationally possible with either the FNM or PNM to replicate the Zipf across all nine orders of magnitude,  as the modeled corpus size $N_t$ grows exponentially (Fig~\ref{Zipf}d).

  \begin{figure}[!h]
  \begin{center}
       \includegraphics[width=2.6in]{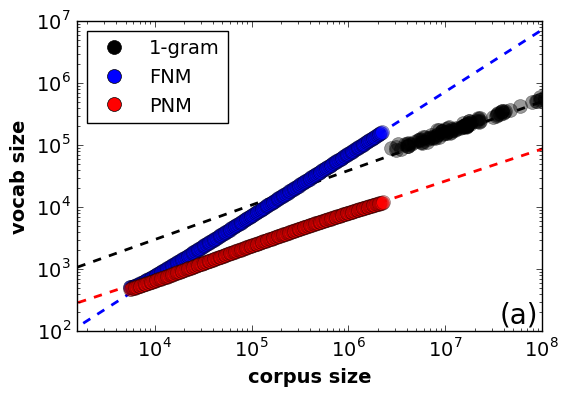}
        \includegraphics[width=2.6in]{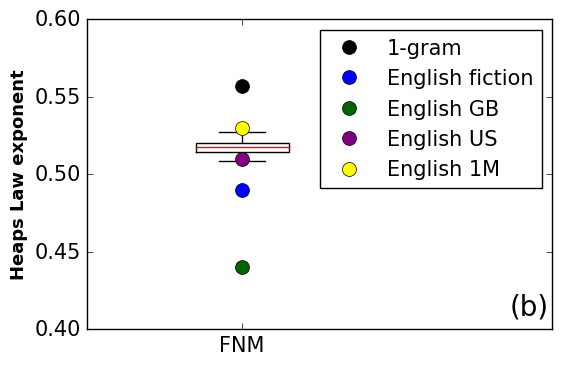}
       \end{center}
 \caption{{\bf Heaps law in simulated Neutral models versus 1-gram data. (a):} A double-logarithmic plot, showing corpus size versus vocabulary size, i.e. Heaps Law, for all 1-grams  (black), the FNM (blue) and the PNM (red). {\bf (b):} The Heaps law exponents, $\beta$, for the data series on the left, as well as additional data series, using Table 1 in \cite{Petersen_etal_2012}: all English 1-grams: $0.54 \pm 0.01$; English fiction:  $0.49 \pm 0.01$; English GB:  $0.44 \pm 0.01$; English US:  $0.51 \pm 0.01$. The 100 independent runs of each neutral model, using parameters listed in the text, yielded  $\beta = 0.52 \pm 0.07$ for the PNM, and $\beta = 1.00 \pm 0.002$ for the FNM (not shown).}
\label{Heaps} 
\end{figure} 

   \begin{figure}[!h]
  \begin{center}
      \includegraphics[width=2.3in]{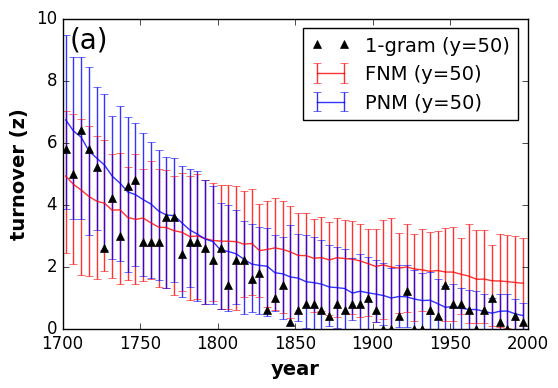}
       \includegraphics[width=2.3in]{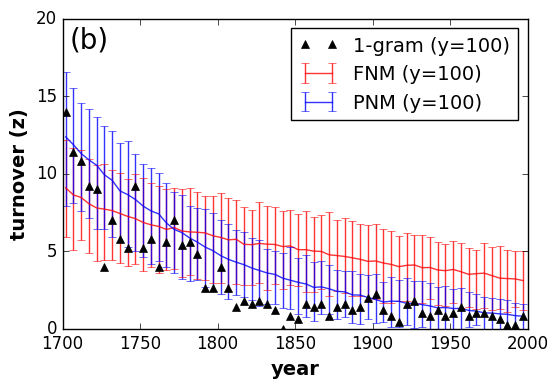}
       \includegraphics[width=2.3in]{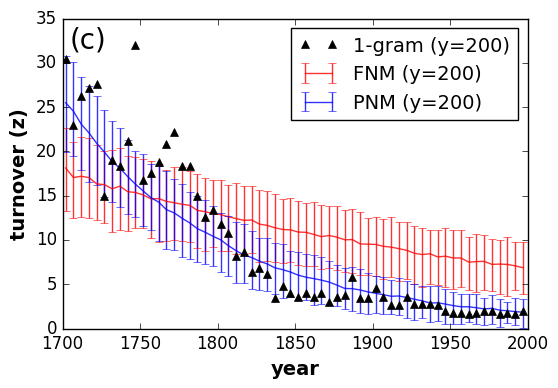}
  \end{center}
 \caption{{\bf Turnover decay in neutral model versus 1-gram data}, for different toplist sizes. Each panel shows the annual turnover among the ranked lists of the top $y$ most frequently-used 1-grams, for list sizes of {\bf(a)}$y=50$, {\bf(b)} $y=100$ and {\bf(c)} $y=200$. The respective line and error bars in each color represent the range of FNM and PNM simulation results. Bands indicate 95\% range of simulated values.}
\label{Turnover_decay} 
\end{figure}

  \begin{figure}[!h]
  \begin{center}
         \includegraphics[width=2.3in,scale=0.5]{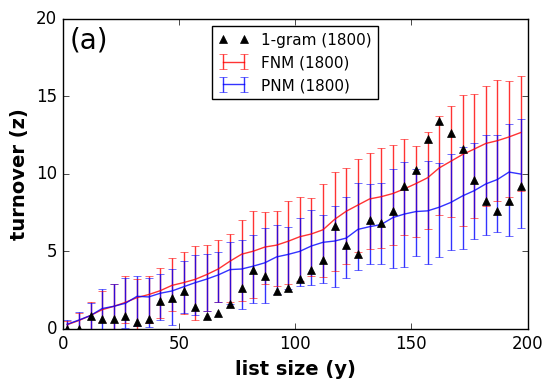}
         \includegraphics[width=2.3in,scale=0.5]{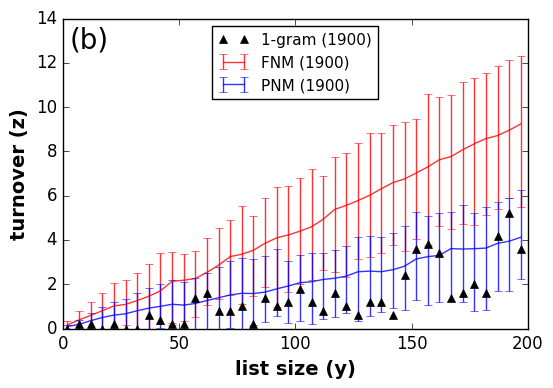}
         \includegraphics[width=2.3in,scale=0.5]{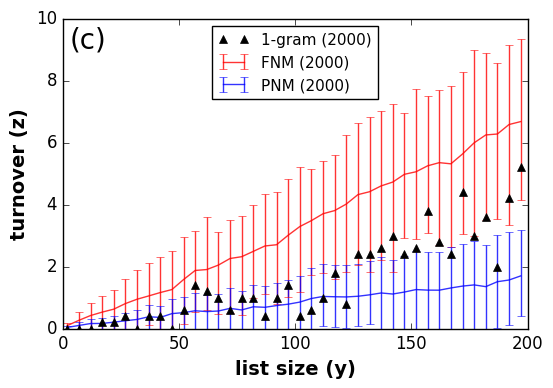}
  \end{center}
 \caption{{\bf Turnover profiles in 1-gram data and in simulated results}, for {\bf(a)} the year 1800, {\bf(b)} the year 1900 and {\bf(c)} the year 2000. In each panel, the circles show turnover $z$ in 1-grams versus list size $y$, averaged over the decade from five years before the new century to five years after. For the FNM, the  corpus size, $N_t$, is 1.5 million by year 2000. For the PNM, the sample $S_t$ grows exponentially as $S_0e^{\alpha t}$ and the sampled {\it canon} size, $N_t$, grows linearly at 10,000 words per year, reaching 3 million by year 2000 ($t=301$). For the FNM and the PNM, bands indicate 95\% range of simulated values.}
\label{Turnover_profile} 
\end{figure}

Fig~\ref{Heaps}a illustrates the relationship between corpus size and vocabulary size in our partial-sampling Neutral model. Due to the exponentially increasing sample size, the ratio of vocabulary size over corpus size becomes increasingly small, thus the model gives us the sub-linear relationship described by $v_t=N_t^\beta$, where $\beta < 1$. On the double-logarithmic plot in Fig~\ref{Heaps}a, the Heaps law exponent is equivalent to the slope of the data series. The PNM matches the 1-gram data with Heaps exponent (slope) of about 0.5, whereas the FNM, with exponent about 1.0, does not. Fig~\ref{Heaps}b shows how 100 runs of the PNM yields a Heaps law exponent within the range derived by \cite{Petersen_etal_2012} for several different n-grams corpora (all English, English fiction, English GB, English US and English 1M).  We also The PNM yields Heaps law exponent $\beta \approx 0.52 \pm 0.006$, within the range of English corpora, whereas the FNM yields a mismatch with the data of $\beta \approx 1 \pm 0.002$ (Fig~\ref{Heaps}b).

In Fig~\ref{Heaps}a, there is a constant offset on the y-axis between vocabulary size in the PNM ($\alpha = 0.02$, $N=10000$) versus the 1-gram data. Both data series follow Heaps exponent $b \approx 0.5$, but the coefficient, $A$, is several times larger for the 1-gram data than for the PNM. We do not think this is due to our choice of canon size $N$ in the PNM, because if we halve it to 5000, the resulting $A$ does not significantly change. The difference could be resolved, however, with larger exponential growth in PNM corpus size, $S_t$, over the 300 time steps. Computationally, we could only model the PNM with growth exponent $\alpha = 0.02$\textemdash using $\alpha = 0.03$, as would fit the actual growth of the n-gram corpus over 300 years \cite{Bentley_etal_2012}, makes the PNM too large to compute. Nevertheless, we can roughly estimate the effect; when we reduce $\alpha$ from 0.02 to 0.01, while keeping $N = 10000$, we find that $A$ averaged over one hundred PNM runs is reduced from $6.3 \pm 0.5$ to $1.4 \pm 0.3$.  Given an exponential relationship, increasing alpha to 0.03 would increase $A$ to about 20, which is within the magnitude of offset we see in Fig~\ref{Heaps}a. Of course, this question can be resolved precisely when the much larger PNM can be simulated.

Regarding dynamic turnover, we consider turnover in ranked lists of size $y$, varying the list size $y$ from the top 1000 most common words down to the top 10 (the top 1 word has been ``the'' since before the year 1700).  We measure turnover in the word-frequency rankings by determining the top $y$ rankings independently for each year, and then counting the number of new words to appear on the list from one year to the next. Fig~\ref{Turnover_decay} shows the number of 1-grams to drop out of the top 1000, top 500 and top 200 per year in the 1-gram data. Annual turnover among the top 1000 and the top 500 decreased exponentially from the year 1700 to 2000, proportional to $e^{-0.012t}$  ($r^2>0.91$ for both), where $t$ is years since 1700. This exponential decay equates to roughly a halving of turnover per century. 

Since the corpus size was increasing with time, Fig~\ref{Turnover_decay} effectively also shows how turnover in top $y$ list decreases as corpus size increases in the partial-sampling Neutral model, where the corpus size grows faster than the number relative to speakers over the years. The exponential decay in turnover in  the partial-sampling Neutral model is markedly different than the base Neutral model, in which turnover would be growing as corpus size grew, due to term $n_s^{0.013}$ in equation~\ref{EG1}. 

Finally, we also look at the ``turnover profile'', plotting list size $y$ versus turnover $z_y$ for different time slices (Fig~\ref{Turnover_profile}). For all words, $z_y \propto y^{1.26}$ for different time periods (Fig~\ref{Turnover_profile}).  We can then compare the turnover profile for the 1-grams to the prediction from eq. \ref{EG1} that turnover will be proportional to $y^{0.86}$, as shown in Fig~\ref{Turnover_profile}b.
  
Table \ref{Tab2} lays out the specific predictions of each of the models and how they fare against empirical data. Bands indicate 95\% range of simulated values. While the predictions for the FNM and PNM are similar for $y = 50$ and for the year 1800 (Fig~\ref{Turnover_decay}a and Fig~\ref{Turnover_profile}a), they do differ substantially in their predictions for Zipf's law and Heaps law under list size $y =200$ and for the year 2000  (Fig~\ref{Turnover_decay}c and Fig~\ref{Turnover_profile}c).  Although the FNM can fit Zipf's Law with the right parameters, it cannot also fit Heaps law or the turnover patterns at the same time as matching Zipf's Law. In contrast, the PNM can fit Zipf's law, Heaps law exponent (Fig~\ref{Heaps}a), and the 2000 series in Fig~\ref{Turnover_decay} (but starts to breakdown at $y > 150$).  Neither the FNM nor the PNM does very well at $y = 200$.

  \begin{table}
  \small
\caption{Seven predictions of the Full Neutral Model (FNM) and Partial Sampling Neutral Model (PNM) and how they fare against 1-gram data.}
\begin{tabular}{@{\vrule height 8pt depth3pt} c|c|c|c|c|c|c|c}
   &  Zipf's & Heaps & Heaps & Turnover & Turnover & $z$ vs $y$ & $z$ vs $y$ \\ 
 Model &  Law & exponent &  coefficient & $y = 50$ &  $y = 200$ & yr 1800 & yr 2000\\ 
 \hline
FNM & {\color{blue} Yes}/No & No & No & {\color{blue} Yes} & No & {\color{blue} Yes} & No\\
PNM & {\color{blue} Yes} & {\color{blue} Yes} & No & {\color{blue} Yes} & {\color{blue} Yes?} & {\color{blue} Yes} & {\color{blue} Yes}
\label{Tab2}
\end{tabular}
\end{table}

\section*{Discussion}

We have explored how `neutral' models of word choice could replicate a series of static and dynamic observations from a historical 1-gram corpora: corpus size, frequency distributions, and turnover within those frequency distributions.  Our goal was to capture two static and three dynamic properties of word frequency statistics in one model. The static properties are not only the well-known (a) Zipf's law, which a range of proportionate-advantage models can replicate, but also (b) Heaps law. The dynamic properties are (c) the continual turnover in words ranked by popularity, (d) the decline in that turnover rate through time, and (e) the relationship between list size and turnover, which we call the turnover profile. 

We found that, although the full-sample Neutral model (FNM) predicts the Zipf's law in ranked word frequencies, the FNM does not replicate Heaps law between corpus and vocabulary size, or the concavity in the non-linear relationship between list size $y$ and turnover $z_y$, or the slowing of this turnover through time among English words. 

It is notable that we found it impossible to capture all five of these properties at once with the FNM. It was a bit like trying to juggle five balls, as soon as the FNM could replicate some of those properties, it dropped the others. Having explored the FNM under broad range of under a range of parameter combinations, we ultimately determined that it could never replicate all these properties at once. This is mainly because both vocabulary size in the FNM is proportional to corpus size (rather than roughly the square root of corpus size as in Heaps law) and also because turnover in FNM should increase slightly with growing population, not decrease as we see in the 1-gram data over 300 years. Other hypotheses to modify the FNM, such as introducing a conformity bias \cite{Acerbi_Bentley_2014}, can also be ruled out. In the case of conformity bias\textemdash where agents choose high-frequency words with even greater probability than just in proportion to frequency\textemdash both the Zipf law and turnover deteriorate under strong conformity in ways that mis-match with the data. 

What did ultimately work very well was our partial-sampling Neutral model, or PNM (Fig~\ref{Models}b), which models a growing sample from a fixed-sized FNM. Our PNM, which takes exponentially increasing sample sizes from a neutrally evolved latent population,  replicated the Zipf's law, Heaps law, and turnover patterns in the 1-gram data. Although it did not replicate exactly the particular 1-gram corpus we used here, the Heaps law exponent yielded by the PNM does fall within the range\textemdash from 0.44 to 0.54\textemdash observed in different English 1-gram corpora \cite{Petersen_etal_2012}.  Among all features we attempted to replicate, the one mismatch between PNM  and the 1-gram data is that the PNM yielded an order of magnitude fewer vocabulary words for a given corpus size, while increasing with corpus size according to the same Heaps law exponent.  The reason for this mismatch appears to be a computational constraint: we could not run the PNM with exponential growth quite as large as that of the actual 300 years of exponential growth in the real English corpus.

As a heuristic device, we consider the fixed-size FNM to represent a canonical literature, while the growing sample represents the real world of exponentially growing numbers of books published ever year in English.  Of course, the world is not as simple as our model; there is no official fixed canon, that canon does not strictly copy words from the previous year only and there are plenty of words being invented that occur outside this canon. 

Our canonical model of the PNM differs somewhat from the explanation by \cite{Petersen_etal_2012}, in which a ``decreasing marginal need for additional words'' as the corpus grows is underlain by the ``dependency network between the common words ... and their more esoteric counterparts.'' In our PNM representation, there is no network structure between words at all, such as ``inter-word statistical dependencies'' \cite{Piantadosi_etal_2011} or grammar as a hierarchical network structure between words \cite{FerreriCancho_etal_2005}. 

\section*{Conclusion}

Since the PNM performed quite well in replicating multiple static and dynamic statistical properties of 1-grams simultaneously, which the FNM could not do, we find two insights. The first is that the FNM remains a powerful representation of word usage dynamics \cite{Bentley_2008, Reali&Griffiths_2009, Hahn&Bentley_2003, Gleeson_etal_2014, Bentley_etal_2011a, Barucca_etal_2015}, but it may need to be embedded in a larger sampling process in order to represent the world. Case studies where the PNM succeeds and the FNM fails could represent situations where mass attention is focused on a small subset of the cultural variants. The same idea seems appropriate for a digital world, where many cultural choices are pre-sorted in ranked lists \cite{Gleeson_etal_2014}.  In the present century, published books contain only a few percent of the verbiage recorded online, with the volume of digital data doubling about every three years. Centuries of prior evolution in published English word use provides valuable context for future study of this digital transition.

\clearpage

{\small 

\section*{Models and data}

Our aim is to compare key summary statistics from simulated data generated by the hypothetical FNM and PNM processes with summary statistics from Google 1-gram data. See Acknowledgements for data source address and the repository location for the Python code used to generate the FNM and PNM.

\subsection*{Neutral models}
The FNM assumes words in a population at time $t$ are selected at random from the population of books at time $t-1$. The population size $N_t$ increases exponentially, $N_0e^{0.021t}$, through time to simulate the exponentially increasing corpus size observed in the Google n-grams data \cite{Bentley_etal_2012}. We ran a genetic algorithm (described in the Appendix 2) to search the model state space to obtain parameter combinations\textemdash latent corpus size $N_t$, innovation fraction $\mu$ and initial population size $N_0$\textemdash that yielded similar summary statistics to the 1-gram data. With the corpus growth exponent  fixed at 0.021, initial corpus size, $N_0$, was constrained by computational capacity.     

Following the genetic algorithm search, the model was initialized with population size $N_0=3000$ and invention fraction $\mu=0.003$. Once steady state was achieved, we permitted the population size in each successive generation to increase at an exponential growth rate comparable to the average annual growth rate of Google 1-gram data until it finally reached $N_{300}=1.5$ million by time step $t=301$. 

At each time $t$ in the FNM, a new set of $N_t$ words enter the modeled corpus. Each word in the corpus, at time $t$, is either a copy of a word from the previous generation of books, with probability $1-\mu$, or else invented as a new word with probability $\mu$. Each of the copied words is selected from $v_{t-1}$ possible words (the vocabulary in the previous time step), which follow a discrete Zipf's law distribution with the probability a word is selected being proportional to the number of copies the word had in the previous population in time step $t-1$ \cite{Bentley_etal_2014b}.

The PNM, represented schematically in Fig~\ref{Models}, draws an exponentially increasing sample (with replacement) from a latent neutrally-evolving canon. We designate the number of words in the sample as $S_t$, and the cumulative number of words in the canon as $N_t$, which grows by a fixed number of words in each time step. This exponentially increasing sample, $S_0e^{\alpha t}$, has an initial population size $S_0=3000$, growth exponent $\alpha = 0.021$, yielding a final sample size $S_{300}=1.5$ million, matching the FNM. The latent population evolves by the rules of the FNM, but with a constant population size of 10000 for each year $t$ (representing a canonical literature from which the main body of authors sample). The {\it cumulative} canon, $N_t$, thus grows by 10,000 words per year. The partial sample, $S_t$, at time $t$ can copy words from all canonical literature, $N_t$, up to that time step. We set $\mu=0.003$ and run for $t=301$ time steps representing years between 1700 and 2000, which are the same parameters used in the FNM.

\subsection*{1-gram data}
The 1-gram data are available as csv files directly from Google's Ngrams site \cite{Google_2016}. As in a previous study \cite{Acerbi_etal_2013}, we removed 1-grams that are common symbols or numbers, and 1-grams containing the same consonant three or more times consecutively.   As in our other studies \cite{Acerbi_etal_2013, Bentley_etal_2012, Bentley_etal_2014a}, we normalized the count of 1-grams using the yearly occurrences of the most common English word, {\it the}. Although we track 1-grams from the year 1700, for turnover statistics we follow other studies  \cite{Petersen_etal_2012} in being cautious about the n-grams record before the year 1800, due to misspelled words before 1800 that were surely  digital scanning errors related to antique printing styles of that may conflate letters such as `s'  and `f' ({\it e.g., myfelf, yourfelf, provifions, increafe, afked} etc).  The code used for modeling is available at: https://github.com/dr2g08/Neutral-evolution-and-turnover-over-centuries-of-English-word-popularity.

\section*{Acknowledgments}
We thank William Brock for comments on an early draft. RAB thanks the Northwestern Institute on Complex Systems for support as a visiting scholar. DR is supported by a grant from the Hobby School of Public Affairs, University of Houston and also by EPSRC grant to the Bristol Centre for Complexity Sciences (EP/I013717/1). AA was supported by a Royal Society Newton Fellowship at Bristol University entitled "Cultural evolution online"; PG was supported by the Leverhulme Trust grant on ``Tipping Points'' (F/00128/BF) awarded to Durham University.}

\newpage

\paragraph*{Appendix 1}
\label{S1_Appendix}
{\bf Neutral model yields Zipf's law.} Recent analytical results \cite{Strimling_etal_2009} show that the expected number of variants of popularity rank $k$ under the stationary distribution is

 \begin{equation}
f_k=\mu N_t \frac{(1-\mu)^{k-1}}{k}\prod_{i=1}^{k-1}\frac{N_t-i}{N_t-i-1+i\mu}.
\label{eq8}
\end{equation}

\noindent Note from this expression \cite{Strimling_etal_2009}, we can find the ratio of $f_{k+1}/f_k$, which is

 \begin{equation}
\frac{f_{k+1}}{f_k}=\frac{\mu N_t \frac{(1-\mu)^{k}}{k+1}\prod_{i=1}^{k}\frac{N_t-i}{N_t-i-1+i\mu}}{\mu N_t \frac{(1-\mu)^{k-1}}{k}\prod_{i=1}^{k-1}\frac{N_t-i}{N_t-i-1+i\mu}}.
\label{eq8}
\end{equation}

which simplifies to

 \begin{equation}
\frac{f_{k+1}}{f_k}=\frac{k(1-\mu)(N_t-k)}{ (k+1)(N_t-k-1+k\mu)}.
\label{eq9}
\end{equation}

\noindent If $N_t$ is large compared to $k$ and $\mu$ is small, then this simplifies to

 \begin{equation}
\frac{f_{k+1}}{f_k} \approx \frac{k}{k+1},
\label{eq9}
\end{equation}

\noindent which is an expression for Zipf's law, because the ratio of the word frequencies is inversely proportional to the ratio of their ranks.

\paragraph*{Appendix 2.}
{\bf Genetic algorithm.} The PNM has five parameters $N$, $\mu$, $S_0$, $\alpha$ and $T$. The number of time steps, $T$ is fixed at 301 (representing calendar years). The exponential growth rate of the sampled population, $\alpha$, is fixed at 0.02. The other three parameters - initial sampled population size ($S_0$), latent population size, $N$, and innovation rate ($\mu$) - are free. We bound potential values of $N$between 5000 and 30000 and $S_0$ between 1000 and 10000. In both cases the lower bound is chosen to ensure a minimum acceptable vocabulary size is reached and the upper bound is limited by computational constraints. The product $N\mu$ was limited between 5 and 90, as the region in which Neutral model yields a reasonable Zipf's law. For the genetic algorithm, the fitnesses were scored by the following equations and a variable values:
\begin{center}
 \begin{tabular}{||c c c c||} 
 \hline
 Summary statistic & Equation & Target variables \\ [0.5ex] 
 \hline\hline
 Heaps Law & $v=An^b$ & A and b \\ 
 \hline
 Zipf's law & $f \sim k^{- \gamma}$ & $\gamma$  \\
 \hline
 Turnover decay ($y=50$) & $z(50) = z_0e^{-\beta_{50} t}$  & $\beta_{50}$ and $z_0$  \\
 \hline
 Turnover decay ($y=100$) & $z(100) = z_0 e^{-\beta_{100} t}$  & $\beta_{100}$ and $z_0$   \\
 \hline
 Turnover decay ($y=200$) & $z(200) = z_0 e^{-\beta_{200} t}$  & $\beta_{200}$ and $z_0$  \\ [1ex] 
 \hline
\end{tabular}
\end{center}

The PNM parameter combination receives a point when each of the target statistics is approximately the same as the equivalent value from the n-grams data.
The genetic algorithm starts with 100 random parameter combinations then the following steps are repeated until they converge on parameter combinations that maximize fitness scores:

\begin{enumerate}
\item The fittest 20\% from the population is passed to the next generation.
\item The remaining 80\% is populated by recombinations of two randomly selected parents from the fittest 20\% from the previous generation.
\item 15\% of the new agents are subject to random mutation of a single parameter to ensure diversity in the population.
\end{enumerate}

\end{document}